\documentclass{article}

\usepackage{arxiv}

\usepackage[utf8]{inputenc} 
\usepackage[T1]{fontenc}    
\usepackage{hyperref}       
\usepackage{url}            
\usepackage{booktabs}       
\usepackage{amsfonts}       
\usepackage{nicefrac}       
\usepackage{microtype}      
\usepackage{lipsum}
\usepackage{graphicx}
\graphicspath{ {./images/} }

\title{Intelligent Cervical Spine Fracture Detection Using Deep Learning Methods}

\author{
	Reza~Behbahani~Nejad \\
	Faculty of Mechanical Engineering \\ K. N. Toosi University of Technology\\ Tehran, Iran
	\texttt{bnreza982@gmail.com} \\
	\And
	Amir Hossein Komijani \\
	Faculty of Mechanical Engineering\\ University of Tehran \\ Tehran, Iran \\
	\texttt{amir.hossein.komijani2000@gmail.com} \\
	\And
	Esmaeil~Najafi \\
	Faculty of Smart Mechatronics And RoboTics (SMART) \\ Saxion University of Applied Sciences\\ M.H. Tromplaan 28 7513 AB Enschede, The Netherlands
	\texttt{e.najafi@saxion.nl} \\
}

\usepackage{amssymb}
\usepackage{amsmath,amssymb,amsfonts}
\usepackage{algorithmic}
\usepackage{graphicx}
\usepackage{textcomp}
\usepackage{xcolor}
\usepackage{hyperref}
\usepackage{multirow}
\usepackage{rotating}
\usepackage{booktabs}
\usepackage{subfig}
\usepackage{balance}
\usepackage{acronym}
\usepackage{tabularx}
\usepackage{booktabs}
\usepackage{multirow}
\usepackage{array}
\usepackage{multirow}
\usepackage{colortbl}
\usepackage{array}
\usepackage{booktabs}
\usepackage{caption}
\usepackage{subfig}
\usepackage[linesnumbered,ruled,vlined]{algorithm2e}
\usepackage[utf8]{inputenc}
\usepackage[T1]{fontenc}
\DeclareUnicodeCharacter{2264}{\leq}

\graphicspath{{Figures/}}

\newcommand{\reffig}[1]{Figure~\ref{#1}}

\newcommand{\reftab}[1]{Table~\ref{#1}}

\newcolumntype{Y}{>{\centering\arraybackslash}X}
\renewcommand{\arraystretch}{1.5}

\acrodef{CNNs}{Convolutional Neural Networks}

\begin{document}

\maketitle

%
%

%
%
%

\begin{abstract}
Cervical spine fractures constitute a critical medical emergency, with the potential for lifelong paralysis or even fatality if left untreated or undetected. Over time, these fractures can deteriorate without intervention. To address the lack of research on the practical application of deep learning techniques for the detection of spine fractures, this study leverages a dataset containing both cervical spine fractures and non-fractured computed tomography images.
This paper introduces a two-stage pipeline designed to identify the presence of cervical vertebrae in each image slice and pinpoint the location of fractures. In the first stage, a multi-input network, incorporating image and image metadata, is trained. This network is based on the Global Context Vision Transformer, and its performance is benchmarked against popular deep learning image classification model. 
In the second stage, a YOLOv8 model is trained to detect fractures within the images, and its effectiveness is compared to YOLOv5. The obtained results indicate that the proposed algorithm significantly reduces the workload of radiologists and enhances the accuracy of fracture detection.
\end{abstract}


\section{Introduction}
\label{sec:Introduction}

\begin{figure}[!t]
	\centering
	\includegraphics[width=\textwidth]{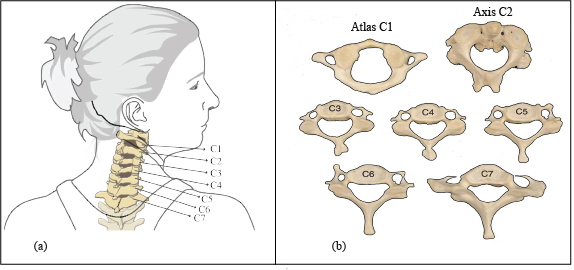}
	\caption{Spine Crevical Veterbraes. Adopted from \cite{neumann2016kinesiology}}
	\label{fig:Spine Crevical Veterbraes}
\end{figure}

The neck is a part of the spinal column, a long, flexible structure that runs through most of the body. The cervical spine, or neck region, is made up of seven bones called vertebrae, which are separated by intervertebral discs as presented in \reffig{fig:Spine Crevical Veterbraes}. The third through the sixth cervical vertebrae show nearly identical features and are therefore considered typical of this region. The upper two cervical vertebrae, the atlas (C1), the axis (C2), and the seventh cervical vertebra (C7) are atypical. Typical Cervical Vertebrae (C3 to C6) have small rectangular bodies made of a relatively dense and strong cortical shell.  the primary function of C1 is to support the head. (C2) has a large, tall body that serves as a base for the upwardly projecting dens. Vertebra Prominens (C7) is the largest of all cervical vertebrae, having many characteristics of thoracic vertebrae \cite{neumann2016kinesiology}.

Cervical Spinal Fractures (CSFx) are serious injuries that can cause death or severe disability due to damage to the spinal cord, the base of the skull where the head meets the spine, and the blood vessels in the neck. If the spinal column is unstable, it can put pressure on the spinal cord and cause further damage \cite{voter2021diagnostic}.

The rate of new spinal cord injuries has remained the same over the past decade, with 26.5 cases per 1 million people \cite{barbiellini2022epidemiology}. Spinal cord injuries are a major cause of disability in young adults and working people, and they have a significant impact on both individuals and society. Therefore, it is important to identify and stabilize CSFx quickly to prevent further disability. Young men are most likely to suffer spinal cord injuries, and the most common causes are traffic accidents, falls, assaults, and sports activities \cite{zanza2023cervical}.

Initial evaluation of a patient with suspected thoracolumbar injury consists of clinical assessment including a thorough neurological examination, and diagnostic imaging. Imaging tests include traditional front-to-back and side-to-side X-rays, Computed Tomography (CT) scans, and Magnetic Resonance Imaging (MRI) scans \cite{galbusera2022biomechanics}.
Detecting and precisely locating vertebral fractures rapidly is imperative to prevent neurological deterioration and paralysis following traumatic incidents. Employing deep learning presents a valuable approach to achieving this objective.

Deep learning (DL) is a type of artificial intelligence (AI) that has been widely used in radiology in recent years. DL has been used to develop automatic fracture detection systems for radiographs of various body parts. With the rise of computer vision models, deep learning, and ubiquitous medical data, it is now possible to develop systems that can assist overworked medical personnel. Quicker diagnosis made possible by DL can prevent lifelong disabilities and even death in some cases. DL can also be used to analyze CT scans, which are difficult for humans to navigate due to the large number of 2D slices \cite{montagnon2020deep}.

Several studies have explored the application of deep learning and computer vision algorithms for detecting cervical spine fractures \cite{chen2019vertebrae, forsberg2017detection}. For instance, a deep convolutional neural network with a bidirectional long-short-term memory (Bi-LSTM) layer for automated fracture detection has been proposed in \cite{salehinejad2021deep} and achieves classification accuracies of 79.18\% on different datasets.  The use of Vision Transformers (ViT) has been explored in\cite{chlkad2023deep} and finds that ViT outperforms traditional \ac{CNNs} with 98\% accuracy in detecting cervical spine fractures. The work \cite{naguib2023classification}focuses on classifying cervical spine injuries as fractures or dislocations using deep learning models, achieving high accuracy, sensitivity, specificity, and precision values. 
The work \cite{paul2023real} utilizing deep learning for automated cervical fracture detection, extensive model optimization through custom layers and data augmentation, and developing a deployable smartphone application.

Additionally, some investigations are primarily focused on the segmentation of vertebrae. They have employed techniques such as U-Net \cite{ronneberger2015u}, either in its 2D form by processing individual slices of spine images as separate inputs to the network \cite{kim2020web, fang2021opportunistic} or in a 3D variant where 3D image patches from multi-slice images serve as input, and a 3D U-Net is trained \cite{lessmann2019iterative, fan2019deep}.

Despite the progress gap persists. While existing methodologies often focus on the segmentation of vertebrae or the detection of fractures, a more holistic approach is warranted—one that not only quantifies the number of cervical vertebrae but also ascertains whether they are fractured or intact. This comprehensive assessment is crucial for providing a thorough clinical evaluation, ensuring that no potential injuries go undetected. Moreover, the dataset employed in the training of deep learning algorithms plays a pivotal role in the accuracy and reliability of fracture detection systems. The quality, diversity, and size of the dataset significantly impact the generalizability and robustness of the algorithms, highlighting the need for standardized, high-quality medical imaging datasets.
The paper is organized as follows: Section \ref{sec:Background} reviews the relevant concepts and materials, and describes the proposed methodology. Section \ref{sec: Results} presents the experimental results, and Section \ref{sec: Discussion} discusses the results and their implications. Finally, Section \ref{sec: Conclusion} summarizes the main contributions of the paper and outlines future work.

\section{Material and methods}
\label{sec:Background}
This section presents an overview of the deep learning models employed in this study for the classification and detection of cervical spine fractures. The subsequent sections of this document provide an in-depth discussion of these concepts.
\subsection{Convolution Neural Networks}
\ac{CNNs} are a type of deep learning model that are specifically designed to process data that has a grid-like topology, such as images. \ac{CNNs} are typically made up of a series of layers, including convolutional layers, pooling layers, and fully connected layers.

The convolutional layer is the most important layer in a CNN architecture. It is responsible for extracting features from the input image by using a series of filters. Each filter is a small matrix of weights that is applied to a small region of the input image. The output of the convolutional layer is a feature map, which is a matrix of values that represent the features that have been extracted from the input image.
The pooling layer is responsible for reducing the spatial size of the feature maps generated by the convolutional layers. This is done by applying a pooling function to each feature map. The pooling function typically takes a small region of the feature map and reduces it to a single value. The most common pooling functions are max pooling and average pooling.
The fully connected layer is the final layer in a CNN architecture. It is responsible for making the final prediction, such as classifying the image or detecting objects in the image. The fully connected layer is a traditional neural network layer, meaning that each neuron in the layer is connected to every neuron in the previous layer \cite{alzubaidi2021review, bhatt2021cnn}.

\ac{CNNs} offer several notable advantages in the realm of image processing. Notably, CNNs incorporate Weight Sharing, Sparse Connectivity, and Local Receptive Fields as integral design principles. Weight Sharing facilitates the sharing of weights across the spatial dimensions of the input image, reducing the number of trainable parameters, thereby enhancing efficiency and mitigating overfitting. Sparse Connectivity ensures that each neuron within a layer maintains connections with only a limited subset of neurons in the preceding layer, further bolstering efficiency and diminishing overfitting concerns. Moreover, \ac{CNNs} employ local receptive fields, restricting each neuron's responsiveness to a small, localized region of the input image, thereby enhancing robustness to noise and variations within the image. These fundamental characteristics collectively contribute to the efficacy of \ac{CNNs} in image analysis and classification tasks \cite{alzubaidi2021review, bhatt2021cnn}.

Various CNN architectures have played pivotal roles in advancing computer vision tasks. VGGNet \cite{simonyan2014very}, ResNet \cite{he2016deep}, DenseNet \cite{huang2017densely}, and ConvNeXt \cite{liu2022convnet} trained on ImageNet's extensive image collection, have consistently outperformed in image classification. Transfer learning, a technique of reusing pre-trained models on new tasks, has profoundly impacted these successes. It provides a solution for effectively training \ac{CNNs} when labeled data is limited, enabling models to leverage prior knowledge acquired during the initial task. This approach not only saves time but also enhances performance by capitalizing on the learned features \cite{bhatt2021cnn}.

Overall \ac{CNNs} have been shown to be very effective at a wide range of computer vision tasks, including image classification, object detection, and image segmentation.

\subsection{Transformers in vision}

\begin{figure}[!t]
	\centering
	\includegraphics[width=\textwidth]{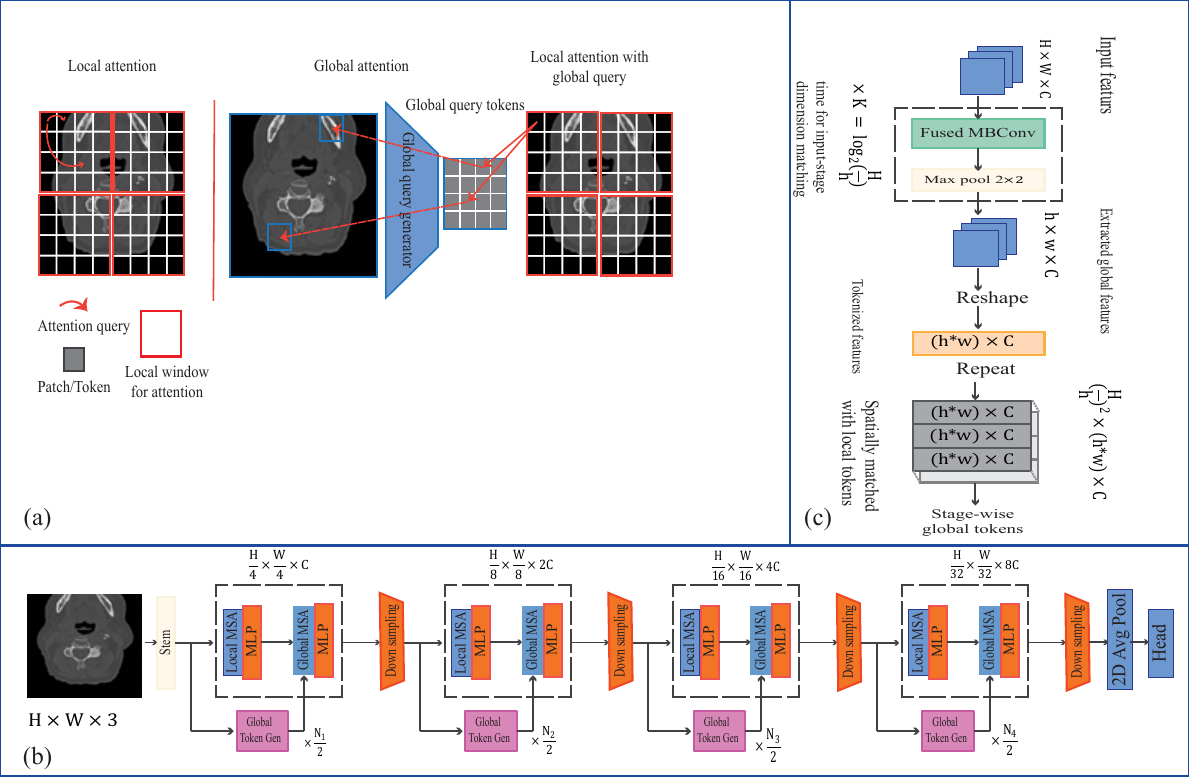}
	\caption{Global Context Vision Transformer network overview for vertebral classification. a. Attention formulation: Local and global attention mechanisms enable cross-region interaction. b. Global query generation: Extraction of global query tokens for long-range information. c. Global query generator: Schematic of the stage-specific dimension transformation and feature extraction process.}
	\label{fig:sys_overview}
\end{figure}
Transformers are prominent deep learning models that have been widely adopted in various fields, such as Natural Language Processing (NLP), Computer Vision (CV), and speech processing. The Transformer architecture represents a major advancement in deep learning, relying solely on attention mechanisms rather than recurrent or convolutional layers for sequence transduction tasks \cite{vaswani2017attention}. The core component is the multi-head self-attention mechanism, which allows the model to capture dependencies between elements in a sequence regardless of position. Multiple parallel self-attention heads compute different transformations of the input to model different types of relationships. Positional encodings are also a critical aspect of injecting order information into the permutation invariant self-attention. Commonly, sine and cosine functions with geometrically increasing wavelengths are used to encode position \cite{khan2022transformers}.

The adoption of transformers in vision comes from the vision transformer model \cite{dosovitskiy2020image}. Vision transformers apply the standard transformer architecture to image classification by converting images to sequences of patches. ViTs work by first dividing an image into a sequence of patches. These patches are then embedded into a high-dimensional space using a linear projection. The embedded patches are then fed into a transformer encoder, which learns to model the relationships between the patches. The output of the transformer encoder is then used to make predictions for the downstream task \cite{khan2022transformers, liu2023survey}.

Although the self-attention mechanism in ViT allows for learning more uniform short and long-range information in comparison to CNN, its monolithic architecture and the quadratic computational complexity of self-attention present hurdles for its rapid implementation in the context of high-resolution images. This is a significant issue, particularly given the critical importance of accurately modeling multi-scale long-range information, as highlighted in \cite{yang2021nvit}.

This led to the development of several architectural variations. The Swin Transformer is a prominent variant that addresses the limitations of the original ViT, particularly its scalability to higher resolution images \cite{liu2021swin}.

The Swin Transformer incorporates a range of innovative features, notably hierarchical feature maps that systematically reduce spatial dimensions, facilitating its adaptability to tasks demanding intricate details. It implements patch merging to aggregate patches into larger tokens, achieving downsampling without reliance on convolutional methods. Moreover, it adopts a window-based multi-head self-attention mechanism, concentrating computational efforts within local windows rather than across all patches for enhanced efficiency. The introduction of shifted windows promotes interactions by creating overlaps between neighboring windows, while orphaned patches are efficiently organized into incomplete windows via cyclic shifting, ensuring global connectivity\cite{liu2021swin}.

Despite the advancements made, self-attention's ability to capture long-range information is challenged by the restricted receptive field of local windows. Additionally, window-connection strategies like shifting only encompass a small surrounding area near each window. 

The Global Context Vision Transformer (GCViT) presents a pioneering vision transformer architecture aimed at optimizing parameter and computational efficiency by simultaneously accounting for local and global spatial interactions within images. GCViT adopts a hierarchical structure characterized by the alternation of local and global self-attention modules as depicted in \reffig{fig:sys_overview}.

In its design, GCViT integrates local self-attention to capture short-range dependencies within defined local windows, a concept akin to prior models such as the Swin Transformer. However, the defining innovation in GCViT is the introduction of global query tokens. These tokens are generated from the entire image through a CNN-like module, and they interact with local key and value tokens within each window. 

This pioneering approach empowers the global self-attention mechanism to effectively capture long-range dependencies spanning across these windows. Remarkably, this is achieved without the need for computationally expensive operations like shifting windows.
Furthermore, GCViT employs a modified convolution block for downsampling, which significantly enhances the modeling of inter-channel dependencies within the network. This distinctive architecture holds immense promise for applications like vertebral classification, offering a powerful blend of local and global information capture efficiently and effectively \cite{hatamizadeh2023global}. 

In general, utilizing transformers in computer vision offers numerous advantages compared to traditional Convolutional Neural Networks (CNNs). ViTs excel in scalability, accommodating larger image sizes while maintaining performance, and exhibit enhanced robustness, displaying resilience to noise and occlusions. Moreover, they stand out for their ease of training, demanding less data for proficient performance. Nevertheless, ViTs come with a few drawbacks, notably the higher computational cost involved in their training and their heightened sensitivity to the size of the training dataset, which surpasses that of CNNs \cite{khan2022transformers}.

\subsection{Object Detection}

Object detection is a key technology in computer vision that identifies and localizes objects of interest within an image or video frame. The task involves both classifying the type of object present as well as determining its spatial location and extent via bounding boxes \cite{alzubaidi2021review}. 

The You Only Look Once (YOLO) \cite{redmon2016you} framework has emerged as a leading approach for real-time object detection. When introduced, YOLO revolutionized object detection with its single neural network architecture that avoids region proposal and sliding window methods used in earlier approaches. The key idea behind YOLO is to divide the input image into a grid and make predictions for bounding boxes and class probabilities in one pass. This unified architecture delivers fast inference speeds while maintaining reasonable accuracy \cite{terven2023comprehensive}.

Since the original YOLO, the architecture has gone through multiple iterations of refinement from YOLOv2 to the latest YOLOv8, gradually improving the accuracy and speed. Key developments include the addition of anchor boxes, improved backbones like DarkNet, multi-scale predictions, and advanced training strategies. While early YOLO models focused on speed, later versions have balanced speed and accuracy by providing lightweight to heavy models\cite{terven2023comprehensive}.

\subsection{Preprocessing}

The dataset used in this work consists of various components, including training and test data, metadata files, and expert annotations.
The ground truth dataset was created by collecting CT imaging data from twelve sites across six continents, encompassing approximately $3,000$ CT studies.  Radiological Society of North America (RSNA) provided expert image-level annotations, indicating the presence, vertebral level, and location of cervical spine fractures. There are target columns, such as patient overall for patient-level outcomes and $C[1-7]$ for individual vertebrae fractures.

\begin{figure}[!t]
	\centering
	\includegraphics[width=\textwidth]{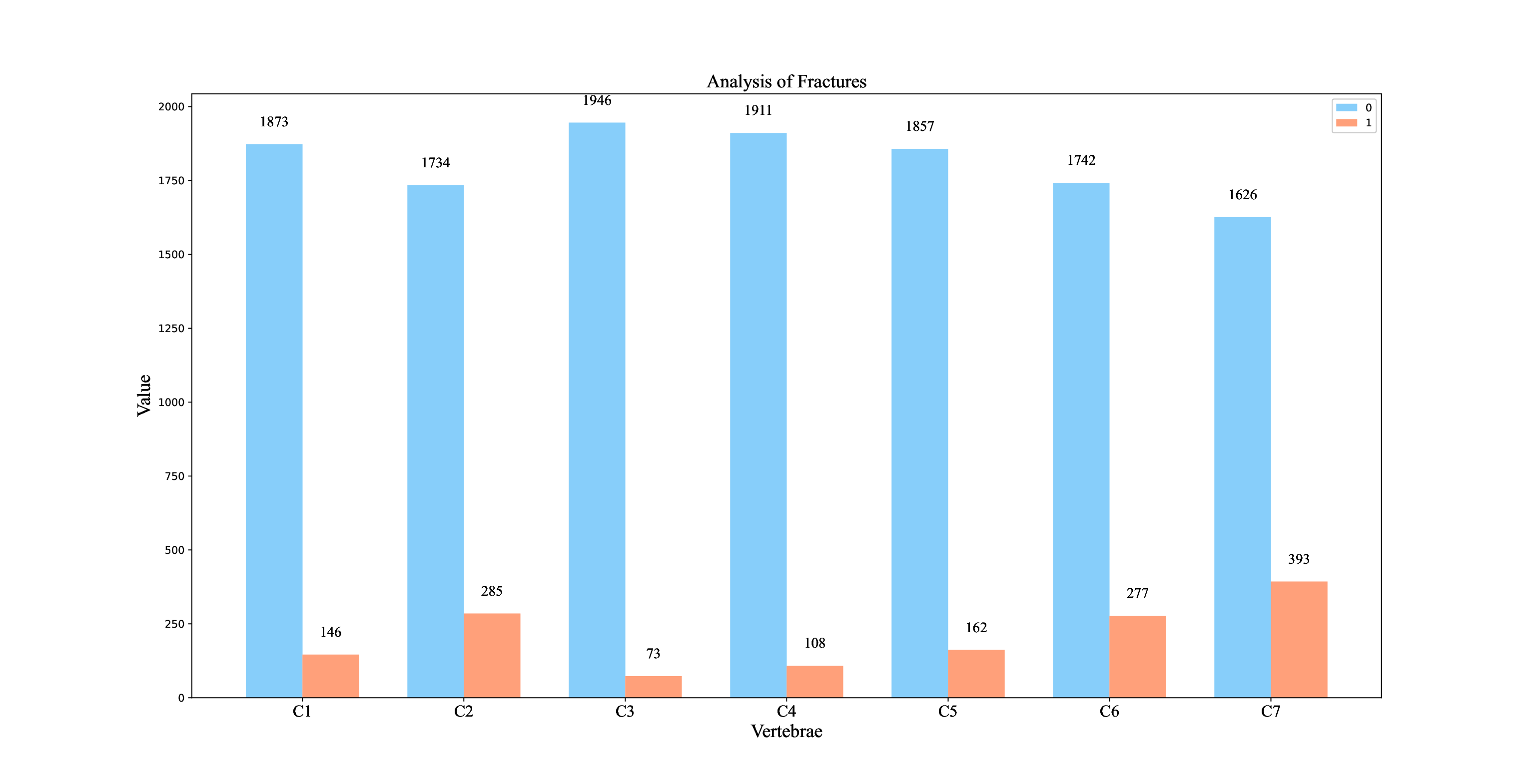}
	\caption{Distribution of fractured and non-fractured cervical vertebrae in the spine}
	\label{fig:Spine_Veterbraes_bar_plot}
\end{figure}

The primary dataset for this task is reasonably balanced, with a split of approximately 52\% for non-fractured and 48\% for fractured cases. Within the fractured cases, there is significant variability, with C7 having the highest proportion of fractures at 19\%, while C3 exhibits the lowest incidence at 4\%. It's noteworthy that some patients may present with multiple fractures, which tend to occur in close proximity, such as between C4 and C5, rather than being dispersed across different vertebrae as shown in \reffig{fig:Spine_Veterbraes_bar_plot}. The medical image data is stored in Digital Imaging and Communications in Medicine (DICOM) format, a well-established standard for medical image storage. Information like image size, pixel dimensions, brightness, contrast, and pixel value range can be extracted from DICOM metadata, providing essential insights for image interpretation.
Additionally, there are  bounding boxes for a subset of the training data.

In addition to DICOM, the dataset also includes Neuroimaging Informatics Technology Initiative (NIfTI) files, a simpler format compared to DICOM, containing the segmentation of vertebrae. However, it's crucial to align the orientation of these segmentations with the DICOM images correctly. The provided segmentations are valuable for locating vertebrae and understanding which vertebrae are present in each image. Additionally, a subset of the dataset includes bounding boxes, which specify the precise location of fractures. These bounding boxes are available for only 12\% of patients in the training set, and it is suggested that training an object localization algorithm could help provide bounding boxes for the entire training set.

\begin{figure}
	\centering
	\subfloat[]{\includegraphics[width=0.22\textwidth]{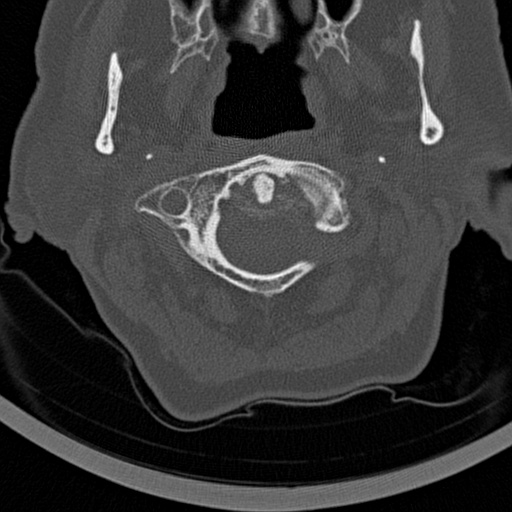}}\hfill
	\subfloat[]{\includegraphics[width=0.22\textwidth]{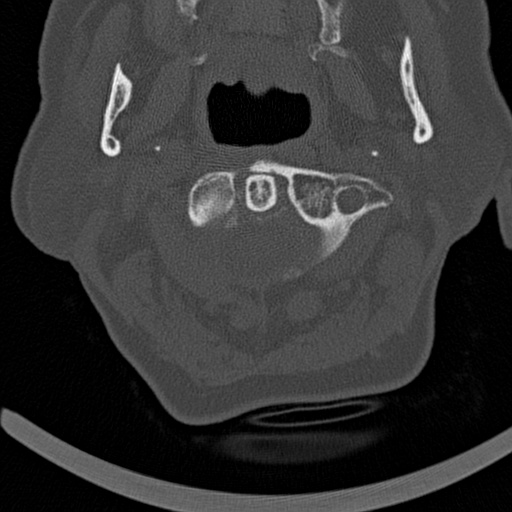}}\hfill
	\subfloat[]{\includegraphics[width=0.22\textwidth]{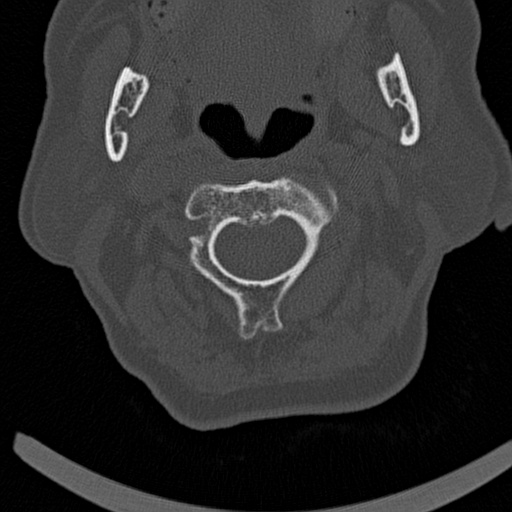}}\hfill
	\subfloat[]{\includegraphics[width=0.22\textwidth]{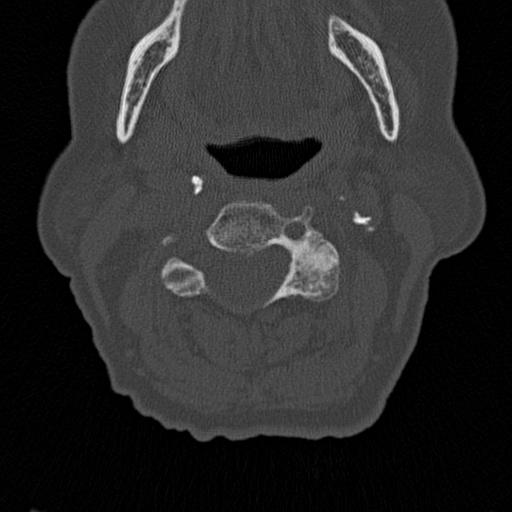}}
	\caption{Visualizing a selection of preprocessed dataset images}
	\label{fig:image1234}
\end{figure}

Overall The preprocessing phase for preparing images to be used in an image classification notebook is showcased by Algorithm 1, with selected outputs depicted in \reffig{fig:image1234}.

\begin{algorithm}[H]
	\SetAlgoNlRelativeSize{0}
	\KwData{DICOM images and NIfTI masks}
	\KwResult{Data preprocessing and custom data generator}
	
	\While{each DICOM image in the dataset}{
		Read the DICOM image\;
		Set the PhotometricInterpretation to YBRFULL\;
		Load the pixel array\;
		Normalize the pixel values\;
		Convert pixel values to a uint8 image\;
		Convert the image to RGB format\;
	}
	
	\While{each NIfTI mask in the dataset}{
		Load the NIfTI mask\;
		Convert the mask to a numpy array\;
		Process the mask: flip, transpose, clip, and convert to uint8\;
		Optionally, apply further transformations to the mask\;
	}
	
	Split the dataset into training and testing sets\

	\caption{Data Preprocessing }
\end{algorithm}

\subsection{Classification of Vertebrae}

\begin{figure}[!t]
	\centering
	\includegraphics[width=\textwidth]{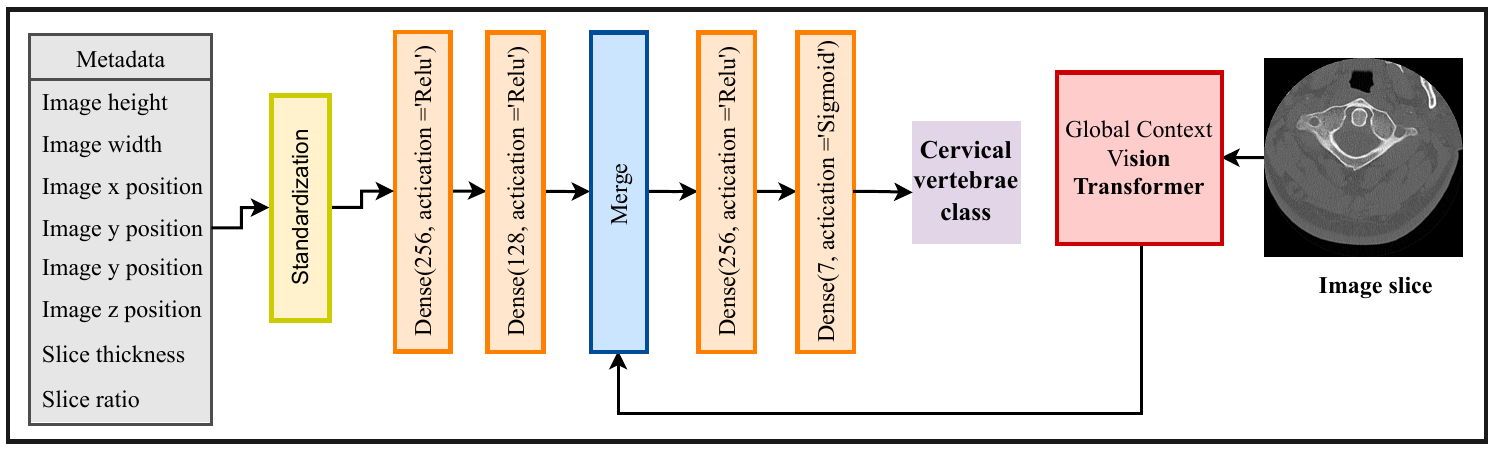}
	\caption{The architecture of the proposed multi-input network. Images are processed through a Global Context Vision Transformer (tiny version), while metadata is input into two fully connected layers.}
	\label{fig:schematic}
\end{figure}

This section outlines the methodology employed for cervical vertebrae classification in the current study. To achieve this objective, a multi-input deep neural network is leveraged for vertebrae classification. This network accommodates two types of inputs: images and metadata, both of which contain essential features such as image positions and slice ratios. Given the sequential nature of image acquisition in CT imaging, the inclusion of these metadata features proves particularly advantageous. The network architecture integrates the Global Context Vision Transformer, with pre-trained weights, to handle the image input, while the metadata input undergoes processing through three fully connected layers. Subsequently, the extracted features from both inputs are concatenated, and two additional fully connected layers are employed for the final classification. The schematic of the network is presented in \reffig{fig:schematic}

Furthermore, a learning rate reduction strategy is implemented through the utilization of the $ReduceLROnPlateau$ callback. This strategy continuously monitors the validation macro F1 score and dynamically adjusts the learning rate during the training process. This adaptive learning rate strategy serves the dual purpose of enhancing model convergence and improving overall performance.

In order to assess the effectiveness of the approach employed, several deep learning models, including ResNet152V2, VGG19, DenseNet, ConvNext, Vision Transformer, and Swin Transformer, are implemented. These models operate solely on image inputs for the classification of vertebrae. The outcomes of these model implementations are summarized in \reftab{tab:cls Performance Metrics}.

\subsection{Fracture Detection}

The dataset used in this study comprises bounding boxes representing the locations of fractures on vertebrae. To enrich the dataset, an additional set of around one thousand 2D images depicting non-fractured vertebrae has been integrated. The inclusion of these non-fractured samples aims to provide a more comprehensive and balanced representation of the data, facilitating improved model training.

In the pursuit of effective fracture detection, this study leverages the capabilities of two object detection algorithms: YOLOv5 and YOLOv8. These algorithms are to identify fracture occurrences within the images. The choice of utilizing YOLOv5 and YOLOv8 is grounded in their proven effectiveness in object detection tasks, making them suitable candidates for the fracture detection objective at hand.

\section{Results}
\label{sec: Results}
The evaluation of the network employs the utilization of multiple metrics in the context of a multi-label classification task. Specifically, the evaluation metrics employed in this study encompass Macro F1, Exact Match Ratio (EMR), and Coverage Error. These metrics are calculated using the Scikit-learn library \cite{scikit-learn}. The Macro F1-score is a metric used for evaluating the performance of a multi-label classification model. It calculates the average F1 score across all the labels, providing a single value that reflects the model's ability to simultaneously balance precision and recall for multiple classes. The F1 score is a harmonic mean of precision and recall, and it takes into account both false positives and false negatives. For each label, it measures how well the model correctly identifies true positives while minimizing false positives and false negatives. The macro F1 score then computes the mean of these label-specific F1 scores, providing a comprehensive evaluation of the model's overall classification performance across all labels. Equation 1 depicts the formula for calculating the Macro F1 score.
\begin{equation}
	\resizebox{\textwidth}{!}{$
		\text{Macro F1} = \frac{1}{N_{\text{LABELS}}} \sum_{i=1}^{N_{\text{LABELS}}} \frac{2 \cdot \text{True Positives}_i}{2 \cdot \text{True Positives}_i + \text{False Negatives}_i + \text{False Positives}_i + \epsilon}
		$}
\end{equation}

Exact Match Ratio (EMR) is another metric used to evaluate the performance of a multi-label classification model. It is calculated by the percentage of instances where the model predicts all of the correct labels for a given instance. EMR is a more stringent metric than MacroF1, as it requires the model to predict all of the correct labels for a given instance in order to be considered correct. The mathematical formula for EMR is expressed as follows in equation 2:
\begin{equation}
	\text{Exact Match Ratio} = \frac{\text{Number of Correctly Predicted Samples}}{\text{Total Number of Samples}}
\end{equation}

Coverage Error is also a metric used to evaluate the performance of a multi-label classification model in terms of its ability to predict at least one of the correct labels for a given instance. It is calculated by the percentage of instances where the model predicts no correct labels for a given instance. Coverage Error is a more lenient metric than EMR, as it allows the model to be partially correct as long as it predicts at least one of the correct labels. The coverage error is calculated as shown in equation 3. 
\begin{equation}
	\text{Coverage Error} = \frac{1}{N} \sum_{i=1}^{N} (\text{Number of Additional Labels}_i)
\end{equation}

\begin{table}[ht!]
	\centering
	\caption{Performance metrics of different cervical spine vertebrae classification models}
	\begin{tabularx}{\textwidth}{>{\hsize=1.4\hsize}X *{5}{>{\hsize=0.92\hsize}X}}
		\toprule
		\textbf{Model} & \textbf{MacroF1} & \textbf{Exact Match Ratio} & \textbf{Coverage Error} & \textbf{Trainable Parameters} & \textbf{Non-trainable Parameters} \\
		\midrule
		Proposed Network & 0.96 & 0.95 & 1.26 & 13,080,663 & 14,683,998 \\
		ViT & 0.94 & 0.90 & 1.41 & 56,937,479 & 30,764,544 \\
		Convext & 0.95 & 0.93 & 1.35 & 9,686,535 & 39,966,816 \\
		InceptionV3 & 0.92 & 0.89 & 1.45 & 526,535 & 21,802,592 \\
		ResNet152V2 & 0.96 & 0.94 & 1.26 & 6,045,703 & 52,812,288 \\
		Swin Transformer & 0.95 & 0.909 & 1.36 & 49,899,081 & 2,768 \\
		\bottomrule
	\end{tabularx}
	\label{tab:cls Performance Metrics}
\end{table}

\begin{figure*}[!t]
	\centering
	\includegraphics[width=1\linewidth]{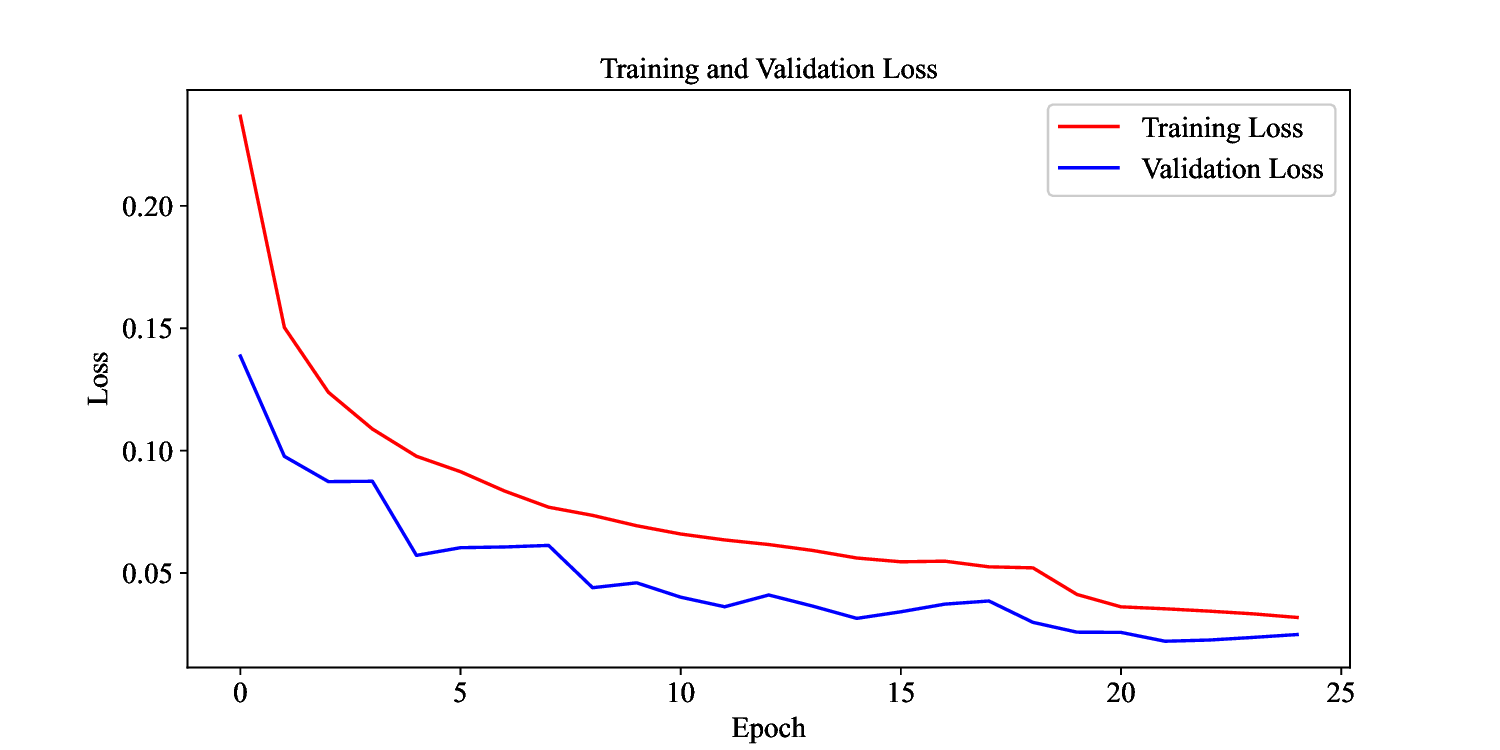}
	\caption{Visualization of Loss for the Multi-Input Network Employing the Global Context Vision Transformer Model in Cervical Vertebrae Classification}
	\label{fig:network Loss}
\end{figure*}

The results of the multi-label classification for cervical spine vertebrae demonstrate the performance of various neural network models. The proposed network, with a MacroF1 score of 0.96 and an Exact Match Ratio of 0.95, exhibits promising results when compared to other models, including ViT, Convext, InceptionV3, ResNet152V2, and Swin Transformer.
\reftab{tab:cls report} displays the classification report for the proposed network, illustrating its strong performance in multi-label cervical spine vertebrae classification. The network achieves high precision (0.97 to 1.00), recall (0.93 to 0.98), and F1-scores (0.95 to 0.99) across all seven classes (C1 to C7), indicating its effectiveness in correctly identifying vertebrae. The micro, macro, and weighted averages are all approximately 0.97, showing consistent overall performance. 
The loss diagram of the proposed network is also presented in \reffig{fig:network Loss}, revealing a consistent downward trend over the course of 25 training epochs.

\begin{table}[ht!]
	\centering
	\caption{Multi-label classification report for cervical spine vertebrae using the proposed network}
	\renewcommand{\arraystretch}{1.5}
	\begin{tabular*}{\textwidth}{@{\extracolsep{\fill}}lcccc}
		\toprule
		\textbf{Class} & \textbf{Precision} & \textbf{Recall} & \textbf{F1-score} & \textbf{Support} \\
		\midrule
		C1 & 0.97 & 0.93 & 0.95 & 284 \\
		C2 & 0.98 & 0.97 & 0.98 & 455 \\
		C3 & 1.00 & 0.98 & 0.99 & 264 \\
		C4 & 0.97 & 0.97 & 0.97 & 272 \\
		C5 & 0.98 & 0.97 & 0.97 & 277 \\
		C6 & 0.99 & 0.96 & 0.97 & 278 \\
		C7 & 0.98 & 0.98 & 0.98 & 320 \\
		\midrule
		Micro avg & 0.98 & 0.97 & 0.97 & 2150 \\
		Macro avg & 0.98 & 0.97 & 0.97 & 2150 \\
		Weighted avg & 0.98 & 0.97 & 0.97 & 2150 \\
		\bottomrule
	\end{tabular*}
	\label{tab:cls report}	
\end{table}

\begin{table}[!t]
	\caption{YOLO Performance Metrics}
	\centering
	\begin{tabularx}{\textwidth}{@{}Xccccccc@{}}
		\toprule
		Model & Class & Images & Instances & Box(P) & R & mAP50 & mAP50-95 \\
		\midrule
		YOLOv8s & all & 1644 & 1444 & 0.94 & 0.921 & 0.96 & 0.747 \\
		& 1 & 1644 & 1444 & 0.94 & 0.921 & 0.96 & 0.747 \\
		\midrule
		YOLOv8m & all & 1644 & 1444 & 0.938 & 0.927 & 0.959 & 0.766 \\
		& 1 & 1644 & 1444 & 0.938 & 0.927 & 0.959 & 0.766 \\
		\midrule
		YOLOv5s & all & 1644 & 1444 & 0.935 & 0.919 & 0.95 & 0.721 \\
		& 1 & 1644 & 1444 & 0.935 & 0.919 & 0.95 & 0.721 \\
		\midrule
		YOLOv5m & all & 1644 & 1444 & 0.932 & 0.931 & 0.958 & 0.748 \\
		& 1 & 1644 & 1444 & 0.932 & 0.931 & 0.958 & 0.748 \\
		\bottomrule
	\end{tabularx}
	\label{tab:YOLO Performance Metrics}
\end{table}

For the cervical vertebrae fracture detection,  the models are trained 100 epochs and the results are demonstrated in \reftab{tab:YOLO Performance Metrics}. 

\begin{figure*}[!t]
	\centering
	\includegraphics[width=1\linewidth]{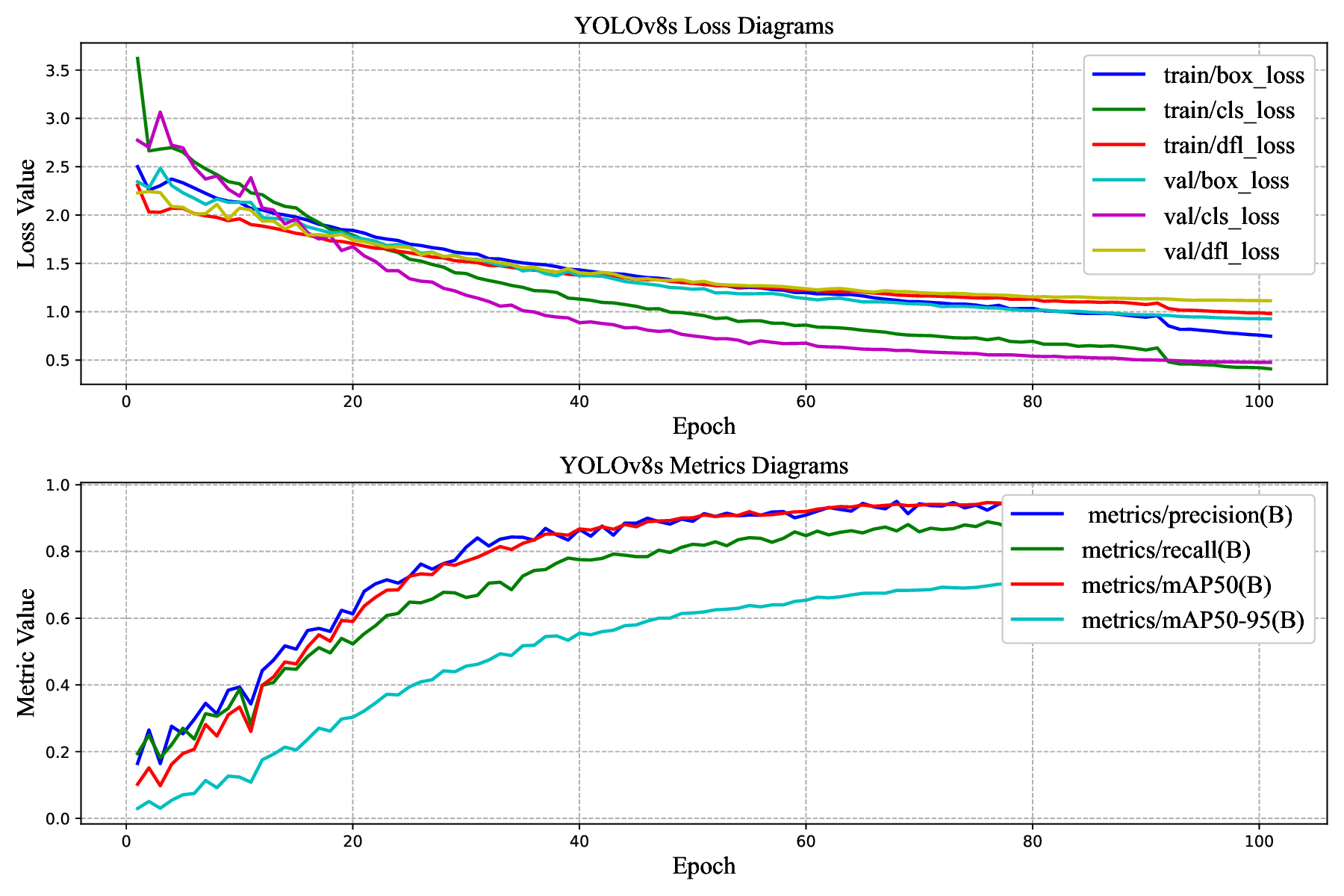}
	\caption{Comprehensive Loss and Metric Analysis for YOLOv8s in Cervical Vertebrae Spine Detection.}
	\label{fig:YOLO8S Loss}
\end{figure*}

\begin{figure*}[!t]
	\centering
	\includegraphics[width=1\linewidth]{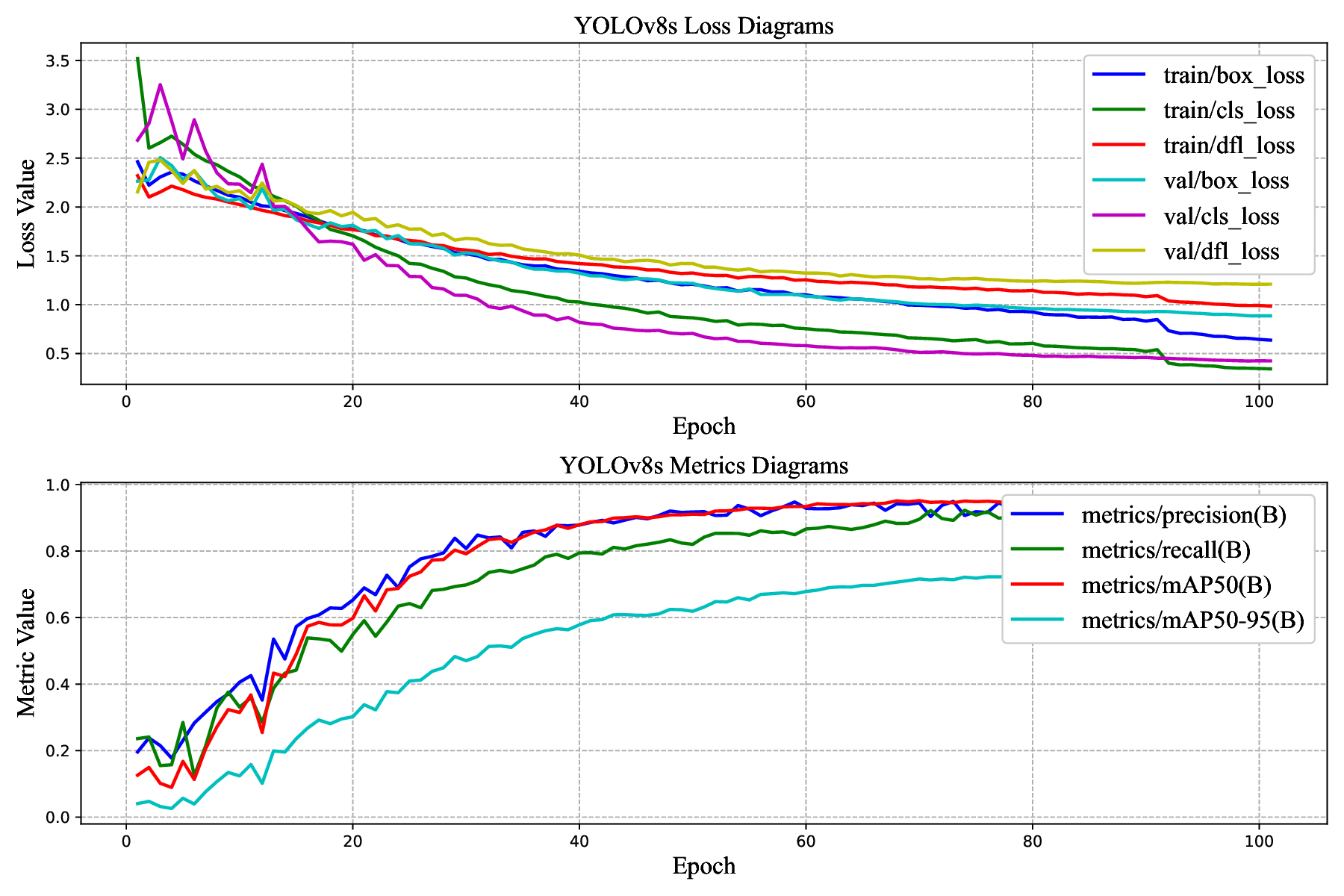}
	\caption{Comprehensive Loss and Metric Analysis for YOLOv8m in Cervical Vertebrae Spine Detection.}
	\label{fig:YOLO8M Loss}
\end{figure*}

mAP50 is the Mean Average Precision (mAP) at an IoU threshold of 0.5. IoU (Intersection over Union) is a metric used to measure the overlap between a predicted bounding box and a ground truth bounding box. A higher IoU threshold means that the predicted bounding box must overlap more with the ground truth bounding box to be considered a true detection.
mAP50-95 is the Mean Average Precision (mAP) at IoU thresholds from 0.5 to 0.95.  This is a more comprehensive metric than mAP50, as it takes into account the model's performance at a wider range of IoU thresholds. Equations 4, 5, 6, and 7 define the formulas for mAP50, mAP50-95, Recall, and Precision.
\begin{equation}
	\text{mAP50} = \frac{1}{N} \sum_{i=1}^{N} \text{AP}_{50}^{(i)}
\end{equation}

\begin{equation}
	\text{mAP50-90} = \frac{1}{N} \sum_{i=1}^{N} \text{AP}_{50-90}^{(i)}
\end{equation}

\begin{equation}
	\text{Recall (R)} = \frac{\text{True Positives}}{\text{True Positives} + \text{False Negatives}}
\end{equation}

\begin{equation}
	\text{Precision (P)} = \frac{\text{True Positives}}{\text{True Positives} + \text{False Positives}}
\end{equation}

\begin{figure}[!t]
	\centering
	\subfloat[]{\includegraphics[width=0.4\textwidth]{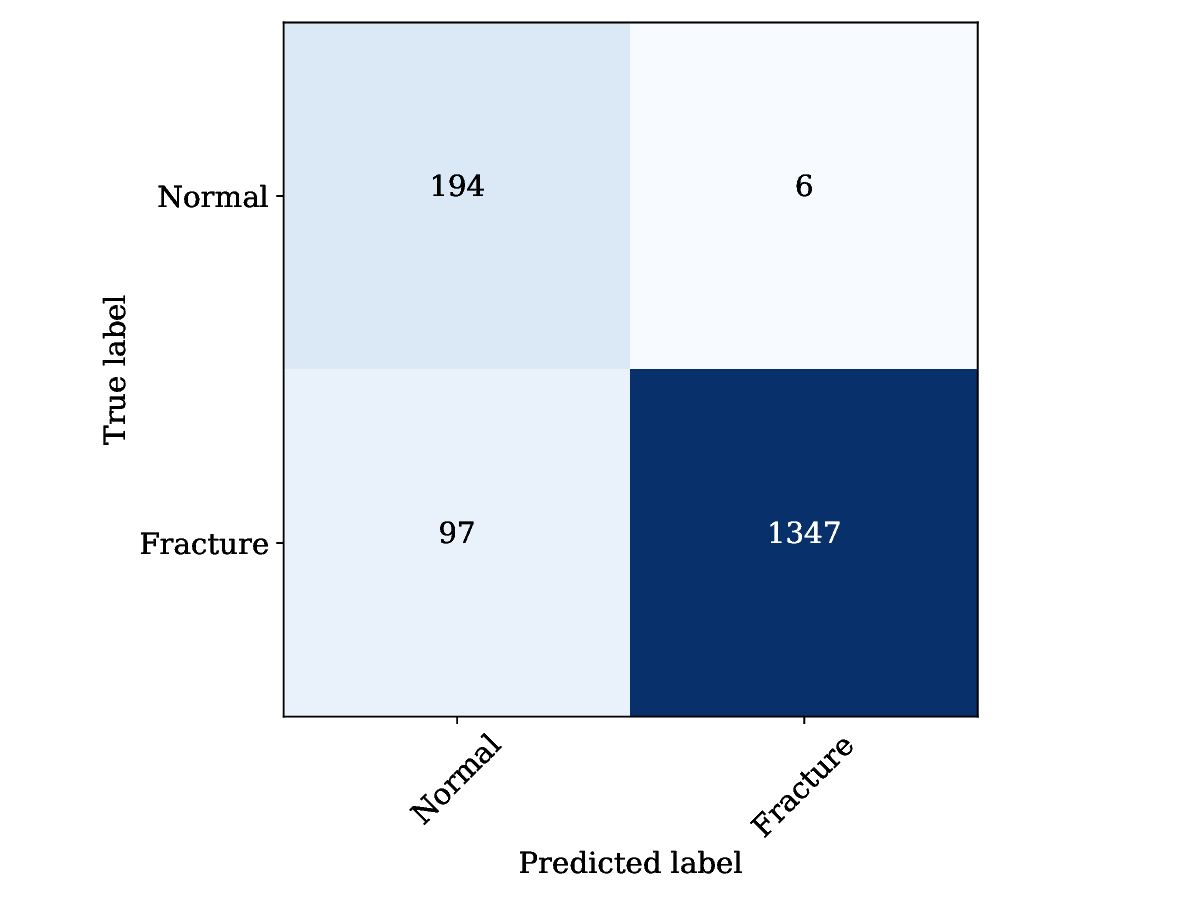}}\hfill
	\subfloat[]{\includegraphics[width=0.4\textwidth]{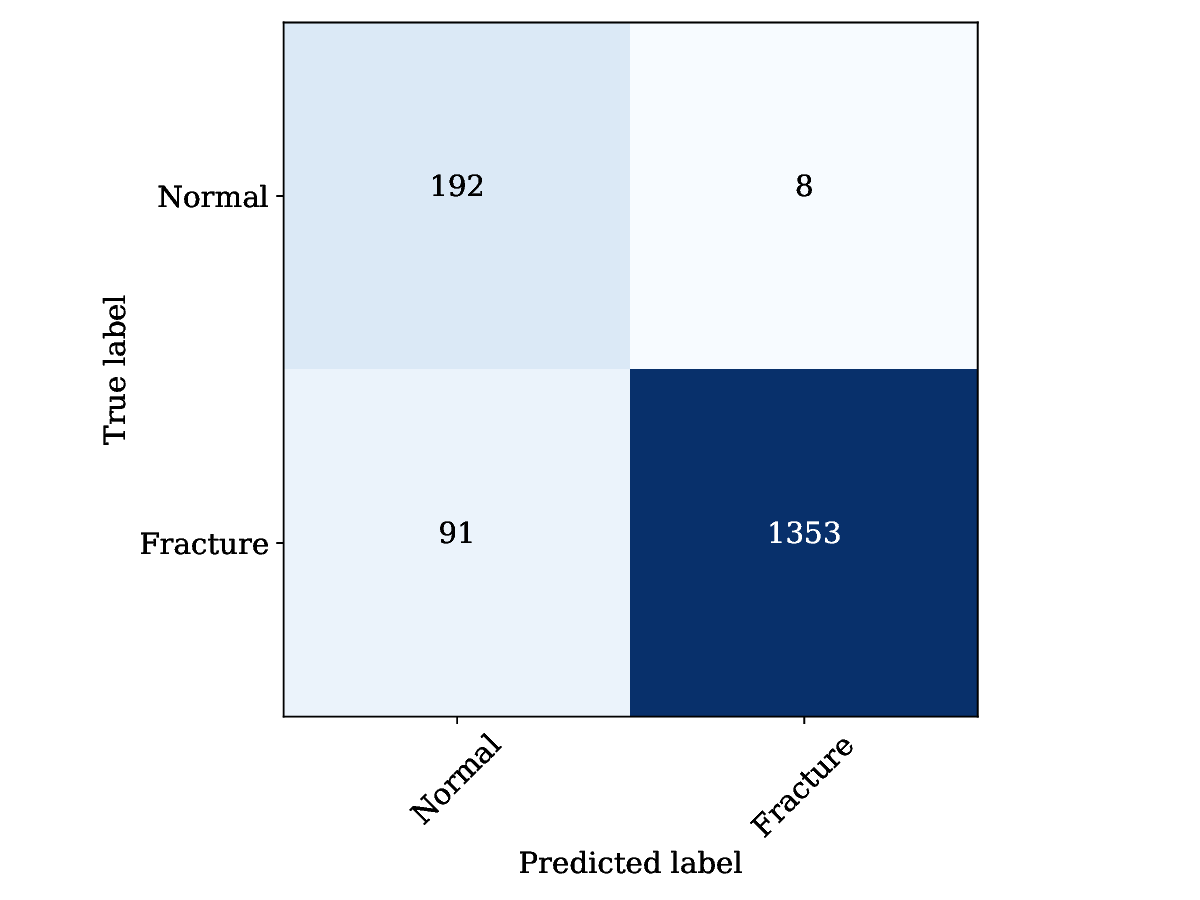}}
	\caption{Confusion Matrices for YOLOv8s and YOLOv8m Models in the Detection of Cervical Vertebrae Fractures}
	\label{fig:yolo_conf}
\end{figure}

\begin{figure}[!t]
	\centering
	\subfloat[]{\includegraphics[width=0.3\textwidth]{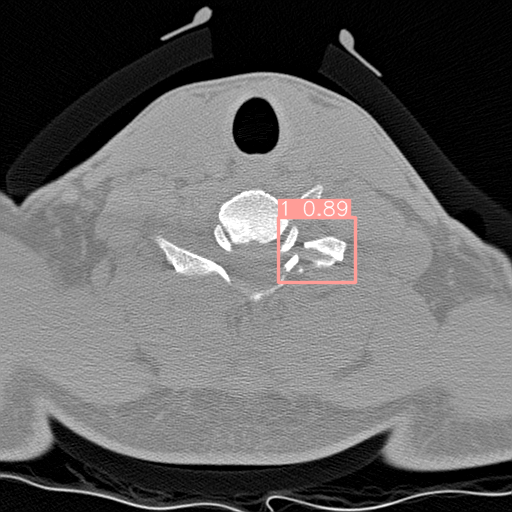}}\hfill
	\subfloat[]{\includegraphics[width=0.3\textwidth]{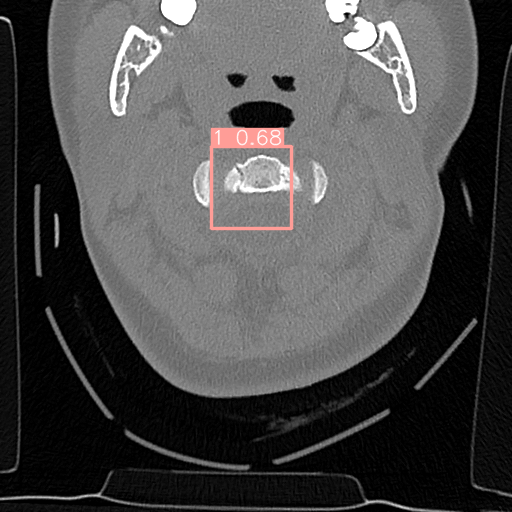}}\hfill
	\subfloat[]{\includegraphics[width=0.3\textwidth]{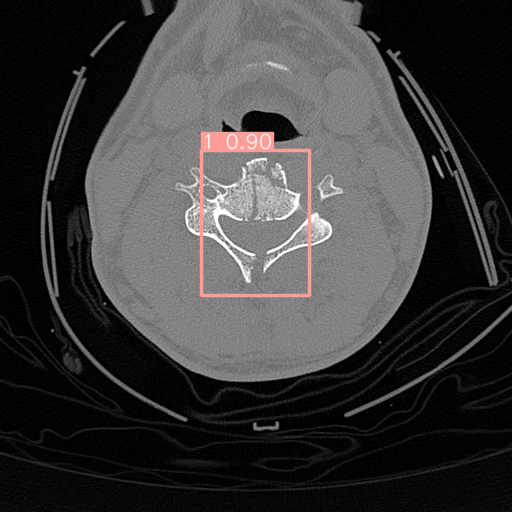}} \\
	\subfloat[]{\includegraphics[width=0.3\textwidth]{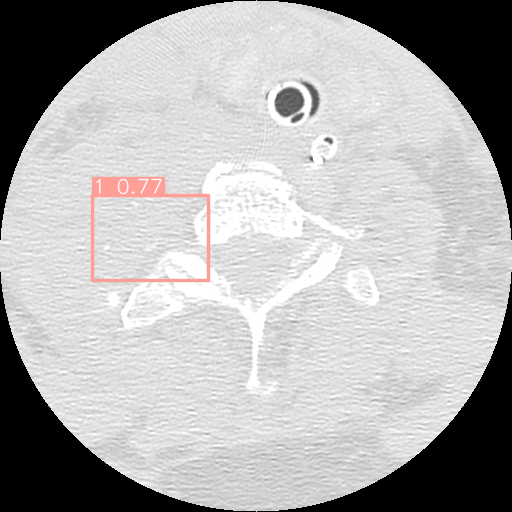}}\hfill
	\subfloat[]{\includegraphics[width=0.3\textwidth]{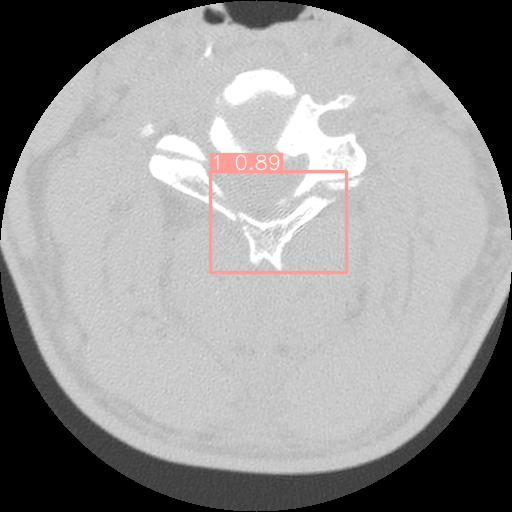}}\hfill
	\subfloat[]{\includegraphics[width=0.3\textwidth]{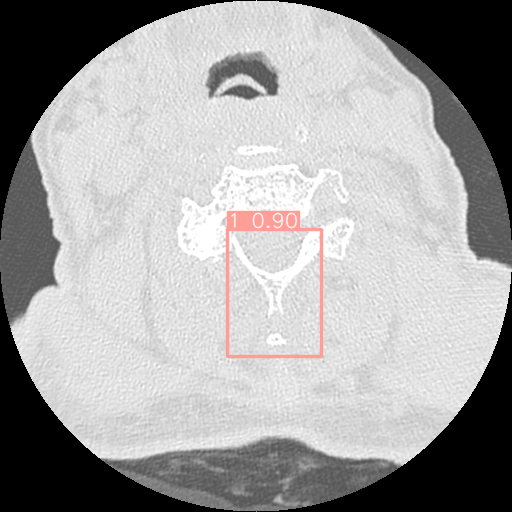}}
	\caption{Visualization of YOLO Algorithm Predictions on Cervical Vertebrae Spine Fracture Detection Dataset Images. Subfigures (a)-(f) display a selection of predictions, showcasing the YOLOv8s model's performance in detecting cervical spine fractures.}
	\label{fig:yolo_predict}
\end{figure}

The loss and metric diagrams for YOLOv8s and YOLOv8m  are also presented in \reffig{fig:YOLO8S Loss} and \reffig{fig:YOLO8M Loss} respectively showing a downward trend.  A curated set of predictions, exemplifying the proficiency of the YOLOv8s model in detecting cervical spine fractures, is visually depicted in Figure \ref{fig:yolo_predict}.

\section{Discussion}
\label{sec: Discussion}
This paper presents a two-step pipeline aimed at detecting cervical vertebrae within individual image slices and locating fractures. In the initial stage, a multi-input network, which includes both the image data and associated image metadata, undergoes training. This network is constructed upon the Global Context Vision Transformer architecture and is evaluated in terms of performance, with a comparison to well-known deep learning image classification models.
In the subsequent stage, a YOLOv8 model is trained specifically for fracture detection in the images, and its performance is assessed in relation to YOLOv5.

One of the notable strengths of the proposed network is its lower Coverage Error, which stands at 1.26. This indicates that it predicts fewer unnecessary labels compared to some of the other models, such as ViT with a Coverage Error of 1.41 and Convext with a Coverage Error of 1.35. This lower Coverage Error suggests that the proposed network produces more precise results, which is a significant advantage for this specific classification task.

On the downside, it's important to consider the relatively high number of non-trainable parameters in the proposed network (14,683,998). High non-trainable parameter counts can lead to increased memory and computational requirements, potentially limiting its practicality in resource-constrained environments.

Overall, the proposed network demonstrates strong performance in cervical spine vertebrae classification, with a competitive MacroF1 score, high Exact Match Ratio, and a notable advantage in terms of Coverage Error. However, the high number of non-trainable parameters is a potential drawback to be addressed for certain deployment scenarios. Careful consideration of the trade-offs between precision and model complexity is crucial when determining the suitability of the proposed network for specific applications.

For fracture detection based on the results, it is clear that the performance of YOLOv8 outperforms YOLOv5. This is a significant improvement, especially considering that YOLOv8 is also faster than YOLOv5. Furthermore, although YOLOv8m has more parameters the mAP50 for YOLOV8s is a bit higher. On the other hand, YOLOv8m has higher mAP50-95. It is evident that YOLOv8s has demonstrated a strong ability to correctly classify images as ``Normal" with 194 true positives and only 6 false negatives as presented in \reffig{fig:yolo_conf}.

However, it tends to make more errors when classifying images as ``Fracture," as indicated by 97 false positives and 1347 true positives. This can be attributed to the nature of medical image analysis, where the cost of missing a ``Fracture" (false negatives) may be considerably higher than misclassifying a ``Normal" image as a ``Fracture" (false positives). YOLOv8m, on the other hand, demonstrates a similar trend but with slightly improved performance when compared to YOLOv8s. It correctly classifies 192 ``Normal" images and 1353 ``Fracture" images. However, it still makes some errors, with 8 false negatives for ``Normal" and 91 false positives for ``Fracture". This model appears to strike a better balance between precision and recall for both classes, indicating a more robust classification performance. 

\section{Conclusion}
\label{sec: Conclusion}
The study introduced a two-stage pipeline leveraging deep learning models for the purpose of classifying vertebrae and detecting cervical spine fractures. The first stage incorporated a multi-input network with a Global Context Vision Transformer (GCViT) for vertebrae classification, while the second stage employed YOLOv8 for fracture detection. The outcomes are subsequently compared against existing deep learning-based image classification models, yielding noteworthy results.
The proposed architecture demonstrated its efficacy, achieving a commendable 96\% Macro F1 accuracy for vertebrae classification and a Mean Average Precision (mAP) of 96\% for fracture detection. 

In terms of future research directions, it is worth exploring the potential integration of segmentation models to further enhance the precision of cervical spine fracture identification. Such segmentation models can facilitate the delineation of distinct anatomical structures within the cervical region, ultimately refining the diagnostic process.

In summary, this study focuses on investigating medical image diagnostics, specifically in the early and accurate identification of cervical spine fractures. The two-stage approach introduced in this research holds promise for improving the management of critical medical injuries and alleviating the burden on radiologists.

\bibliographystyle{unsrt}  
\bibliography{RezaSpine.bib}

\begin{thebibliography}{10}

\bibitem{neumann2016kinesiology}
Donald~A Neumann.
\newblock {\em Kinesiology of the musculoskeletal system-e-book: foundations
  for rehabilitation}.
\newblock Elsevier Health Sciences, 2016.

\bibitem{voter2021diagnostic}
AF~Voter, ME~Larson, JW~Garrett, and J-PJ Yu.
\newblock Diagnostic accuracy and failure mode analysis of a deep learning
  algorithm for the detection of cervical spine fractures.
\newblock {\em American Journal of Neuroradiology}, 42(8):1550--1556, 2021.

\bibitem{barbiellini2022epidemiology}
Claudio Barbiellini~Amidei, Laura Salmaso, Stefania Bellio, and Mario Saia.
\newblock Epidemiology of traumatic spinal cord injury: a large
  population-based study.
\newblock {\em Spinal Cord}, 60(9):812--819, 2022.

\bibitem{zanza2023cervical}
Christian Zanza, Gilda Tornatore, Cristina Naturale, Yaroslava Longhitano,
  Angela Saviano, Andrea Piccioni, Aniello Maiese, Michela Ferrara, Gianpietro
  Volonnino, Giuseppe Bertozzi, et~al.
\newblock Cervical spine injury: Clinical and medico-legal overview.
\newblock {\em La radiologia medica}, 128(1):103--112, 2023.

\bibitem{galbusera2022biomechanics}
Fabio Galbusera.
\newblock Biomechanics of the spine.
\newblock In {\em Human Orthopaedic Biomechanics}, pages 265--283. Elsevier,
  2022.

\bibitem{montagnon2020deep}
Emmanuel Montagnon, Milena Cerny, Alexandre Cadrin-Ch{\^e}nevert, Vincent
  Hamilton, Thomas Derennes, Andr{\'e} Ilinca, Franck Vandenbroucke-Menu, Simon
  Turcotte, Samuel Kadoury, and An~Tang.
\newblock Deep learning workflow in radiology: a primer.
\newblock {\em Insights into imaging}, 11:1--15, 2020.

\bibitem{chen2019vertebrae}
Yizhi Chen, Yunhe Gao, Kang Li, Liang Zhao, and Jun Zhao.
\newblock Vertebrae identification and localization utilizing fully
  convolutional networks and a hidden markov model.
\newblock {\em IEEE Transactions on Medical Imaging}, 39(2):387--399, 2019.

\bibitem{forsberg2017detection}
Daniel Forsberg, Erik Sj{\"o}blom, and Jeffrey~L Sunshine.
\newblock Detection and labeling of vertebrae in mr images using deep learning
  with clinical annotations as training data.
\newblock {\em Journal of digital imaging}, 30(4):406--412, 2017.

\bibitem{salehinejad2021deep}
Hojjat Salehinejad, Edward Ho, Hui-Ming Lin, Priscila Crivellaro, Oleksandra
  Samorodova, Monica~Tafur Arciniegas, Zamir Merali, Suradech Suthiphosuwan,
  Aditya Bharatha, Kristen Yeom, et~al.
\newblock Deep sequential learning for cervical spine fracture detection on
  computed tomography imaging.
\newblock In {\em 2021 IEEE 18th International Symposium on Biomedical Imaging
  (ISBI)}, pages 1911--1914. IEEE, 2021.

\bibitem{chlkad2023deep}
Pawe{\l} Ch{\l}{\k{a}}d and Marek~R Ogiela.
\newblock Deep learning and cloud-based computation for cervical spine fracture
  detection system.
\newblock {\em Electronics}, 12(9):2056, 2023.

\bibitem{naguib2023classification}
Soaad~M Naguib, Hanaa~M Hamza, Khalid~M Hosny, Mohammad~K Saleh, and Mohamed~A
  Kassem.
\newblock Classification of cervical spine fracture and dislocation using
  refined pre-trained deep model and saliency map.
\newblock {\em Diagnostics}, 13(7):1273, 2023.

\bibitem{paul2023real}
Showmick~Guha Paul, Arpa Saha, and Md~Assaduzzaman.
\newblock A real-time deep learning approach for classifying cervical spine
  fractures.
\newblock {\em Healthcare Analytics}, page 100265, 2023.

\bibitem{ronneberger2015u}
Olaf Ronneberger, Philipp Fischer, and Thomas Brox.
\newblock U-net: Convolutional networks for biomedical image segmentation.
\newblock In {\em Medical Image Computing and Computer-Assisted
  Intervention--MICCAI 2015: 18th International Conference, Munich, Germany,
  October 5-9, 2015, Proceedings, Part III 18}, pages 234--241. Springer, 2015.

\bibitem{kim2020web}
Young~Jae Kim, Bilegt Ganbold, and Kwang~Gi Kim.
\newblock Web-based spine segmentation using deep learning in computed
  tomography images.
\newblock {\em Healthcare informatics research}, 26(1):61--67, 2020.

\bibitem{fang2021opportunistic}
Yijie Fang, Wei Li, Xiaojun Chen, Keming Chen, Han Kang, Pengxin Yu, Rongguo
  Zhang, Jianwei Liao, Guobin Hong, and Shaolin Li.
\newblock Opportunistic osteoporosis screening in multi-detector ct images
  using deep convolutional neural networks.
\newblock {\em European Radiology}, 31:1831--1842, 2021.

\bibitem{lessmann2019iterative}
Nikolas Lessmann, Bram Van~Ginneken, Pim~A De~Jong, and Ivana I{\v{s}}gum.
\newblock Iterative fully convolutional neural networks for automatic vertebra
  segmentation and identification.
\newblock {\em Medical image analysis}, 53:142--155, 2019.

\bibitem{fan2019deep}
Guoxin Fan, Huaqing Liu, Zhenhua Wu, Yufeng Li, Chaobo Feng, Dongdong Wang, Jie
  Luo, WM~Wells, and Shisheng He.
\newblock Deep learning--based automatic segmentation of lumbosacral nerves on
  ct for spinal intervention: a translational study.
\newblock {\em American Journal of Neuroradiology}, 40(6):1074--1081, 2019.

\bibitem{alzubaidi2021review}
Laith Alzubaidi, Jinglan Zhang, Amjad~J Humaidi, Ayad Al-Dujaili, Ye~Duan,
  Omran Al-Shamma, Jos{\'e} Santamar{\'\i}a, Mohammed~A Fadhel, Muthana
  Al-Amidie, and Laith Farhan.
\newblock Review of deep learning: Concepts, cnn architectures, challenges,
  applications, future directions.
\newblock {\em Journal of big Data}, 8:1--74, 2021.

\bibitem{bhatt2021cnn}
Dulari Bhatt, Chirag Patel, Hardik Talsania, Jigar Patel, Rasmika Vaghela,
  Sharnil Pandya, Kirit Modi, and Hemant Ghayvat.
\newblock Cnn variants for computer vision: History, architecture, application,
  challenges and future scope.
\newblock {\em Electronics}, 10(20):2470, 2021.

\bibitem{simonyan2014very}
Karen Simonyan and Andrew Zisserman.
\newblock Very deep convolutional networks for large-scale image recognition.
\newblock {\em arXiv preprint arXiv:1409.1556}, 2014.

\bibitem{he2016deep}
Kaiming He, Xiangyu Zhang, Shaoqing Ren, and Jian Sun.
\newblock Deep residual learning for image recognition.
\newblock In {\em Proceedings of the IEEE conference on computer vision and
  pattern recognition}, pages 770--778, 2016.

\bibitem{huang2017densely}
Gao Huang, Zhuang Liu, Laurens Van Der~Maaten, and Kilian~Q Weinberger.
\newblock Densely connected convolutional networks.
\newblock In {\em Proceedings of the IEEE conference on computer vision and
  pattern recognition}, pages 4700--4708, 2017.

\bibitem{liu2022convnet}
Zhuang Liu, Hanzi Mao, Chao-Yuan Wu, Christoph Feichtenhofer, Trevor Darrell,
  and Saining Xie.
\newblock A convnet for the 2020s.
\newblock In {\em Proceedings of the IEEE/CVF conference on computer vision and
  pattern recognition}, pages 11976--11986, 2022.

\bibitem{vaswani2017attention}
Ashish Vaswani, Noam Shazeer, Niki Parmar, Jakob Uszkoreit, Llion Jones,
  Aidan~N Gomez, {\L}ukasz Kaiser, and Illia Polosukhin.
\newblock Attention is all you need.
\newblock {\em Advances in neural information processing systems}, 30, 2017.

\bibitem{khan2022transformers}
Salman Khan, Muzammal Naseer, Munawar Hayat, Syed~Waqas Zamir, Fahad~Shahbaz
  Khan, and Mubarak Shah.
\newblock Transformers in vision: A survey.
\newblock {\em ACM computing surveys (CSUR)}, 54(10s):1--41, 2022.

\bibitem{dosovitskiy2020image}
Alexey Dosovitskiy, Lucas Beyer, Alexander Kolesnikov, Dirk Weissenborn,
  Xiaohua Zhai, Thomas Unterthiner, Mostafa Dehghani, Matthias Minderer, Georg
  Heigold, Sylvain Gelly, et~al.
\newblock An image is worth 16x16 words: Transformers for image recognition at
  scale.
\newblock {\em arXiv preprint arXiv:2010.11929}, 2020.

\bibitem{liu2023survey}
Yang Liu, Yao Zhang, Yixin Wang, Feng Hou, Jin Yuan, Jiang Tian, Yang Zhang,
  Zhongchao Shi, Jianping Fan, and Zhiqiang He.
\newblock A survey of visual transformers.
\newblock {\em IEEE Transactions on Neural Networks and Learning Systems},
  2023.

\bibitem{yang2021nvit}
Huanrui Yang, Hongxu Yin, Pavlo Molchanov, Hai Li, and Jan Kautz.
\newblock Nvit: Vision transformer compression and parameter redistribution.
\newblock 2021.

\bibitem{liu2021swin}
Ze~Liu, Yutong Lin, Yue Cao, Han Hu, Yixuan Wei, Zheng Zhang, Stephen Lin, and
  Baining Guo.
\newblock Swin transformer: Hierarchical vision transformer using shifted
  windows.
\newblock In {\em Proceedings of the IEEE/CVF international conference on
  computer vision}, pages 10012--10022, 2021.

\bibitem{hatamizadeh2023global}
Ali Hatamizadeh, Hongxu Yin, Greg Heinrich, Jan Kautz, and Pavlo Molchanov.
\newblock Global context vision transformers.
\newblock In {\em International Conference on Machine Learning}, pages
  12633--12646. PMLR, 2023.

\bibitem{redmon2016you}
Joseph Redmon, Santosh Divvala, Ross Girshick, and Ali Farhadi.
\newblock You only look once: Unified, real-time object detection.
\newblock In {\em Proceedings of the IEEE conference on computer vision and
  pattern recognition}, pages 779--788, 2016.

\bibitem{terven2023comprehensive}
Juan Terven and Diana Cordova-Esparza.
\newblock A comprehensive review of yolo: From yolov1 to yolov8 and beyond.
\newblock {\em arXiv preprint arXiv:2304.00501}, 2023.

\bibitem{scikit-learn}
F.~Pedregosa, G.~Varoquaux, A.~Gramfort, V.~Michel, B.~Thirion, O.~Grisel,
  M.~Blondel, P.~Prettenhofer, R.~Weiss, V.~Dubourg, J.~Vanderplas, A.~Passos,
  D.~Cournapeau, M.~Brucher, M.~Perrot, and E.~Duchesnay.
\newblock Scikit-learn: Machine learning in {P}ython.
\newblock {\em Journal of Machine Learning Research}, 12:2825--2830, 2011.

\end{thebibliography}

\end{document}